\pdfoutput=1

\documentclass[11pt]{article}

\usepackage{ACL2023}

\usepackage{times}
\usepackage{latexsym}
\usepackage{graphicx}
\usepackage{caption}
\usepackage{graphicx}
\usepackage{multirow}
\usepackage{amsmath}
\usepackage{subcaption}
\usepackage{algorithm}
\usepackage{algorithmic}
\usepackage{tabularx}
\usepackage{booktabs}
\usepackage{newfloat}
\usepackage{adjustbox}
\usepackage{xcolor}

\usepackage[T1]{fontenc}

\usepackage[utf8]{inputenc}

\usepackage{microtype}

\usepackage{inconsolata}

\title{Non-Parametric Temporal Adaptation for Hashtag Prediction}

\author{
Fatemehsadat Mireshghallah\textsuperscript{\rm 1 $\ast$},
Nikolai Vogler\textsuperscript{\rm 1 $\ast$},
Junxian He\textsuperscript{\rm 2} \\
\textbf{Omar Florez}\textsuperscript{\rm 3},
\textbf{Ahmed El-Kishky}\textsuperscript{\rm 3},
\textbf{Taylor Berg-Kirkpatrick}\textsuperscript{\rm 1,3} \\
    \textsuperscript{\rm 1}University of California San Diego,
    \textsuperscript{\rm 2}Carnegie Mellon University,
    \textsuperscript{\rm 3}Twitter Cortex\\
    \{fmireshg,nvogler\}@ucsd.edu, junxianh@cs.cmu.edu,\{oflorez,aelkishky,taylorb\}@twitter.com \\
}

\begin{document}
\maketitle
\begin{abstract}
User-generated social media data is constantly changing as new trends influence online discussion and personal information is deleted due to privacy concerns.
However, most current NLP models are static and rely on fixed training data, which means they are unable to adapt to temporal change---both test distribution shift and deleted training data---without frequent, costly re-training.
In this paper, we study temporal adaptation through the task of longitudinal hashtag prediction and propose a non-parametric dense retrieval technique, which does not require re-training, as a simple but effective solution.
In experiments on a newly collected, publicly available, year-long Twitter dataset exhibiting temporal distribution shift, our method improves by $64.12\%$ over the best parametric baseline without  any of its costly gradient-based updating.
Our dense retrieval approach is also particularly well-suited to dynamically deleted user data in line with data privacy laws, with negligible computational cost and performance loss.

\end{abstract}

%
%
%
%
%

\section{Introduction}
Distribution shift, particularly in the target labels of classification tasks, presents a serious challenge for the deployment of NLP systems in real-world scenarios. 
The distribution of text data changes  over time due to new language usage, events, or  trends \citep{eisenstein2014diffusion,ryskina2020new,jaidka-etal-2018-diachronic}, which causes statically-trained models to become stale without further re-training~\citep{lazaridou2021mind,dhingra2022time,luu-etal-2022-time}. 
Diachronic drift is most evident in social media data, which we study through multi-label tweet-hashtag prediction, where tweets are classified into rapidly changing user-generated trends.
Evidently, recent temporal analysis of named entity recognition models for social media \citep{rijhwani-preotiuc-pietro-2020-temporally} identifies the advantages of continual exposure to late-breaking training data.
Moreover, NLP systems must accommodate deletions as user-generated training data must be removed from platforms for a variety of reasons, e.g., hate speech, disinformation, or obliging user-initiated deletions. 
Given the ever-shifting nature of user-generated data, deployed systems need to be designed to allow them to (1) \textit{adapt} to dynamic test distributions to prevent temporal performance degradation, and (2) \textit{abide} by data privacy laws, such as GDPR, CCPA, etc.~\cite{voigt2017eu}, that
mandate prompt removal of user data.

\begin{figure}[t!]
    \centering
     \includegraphics[width=0.92\linewidth]{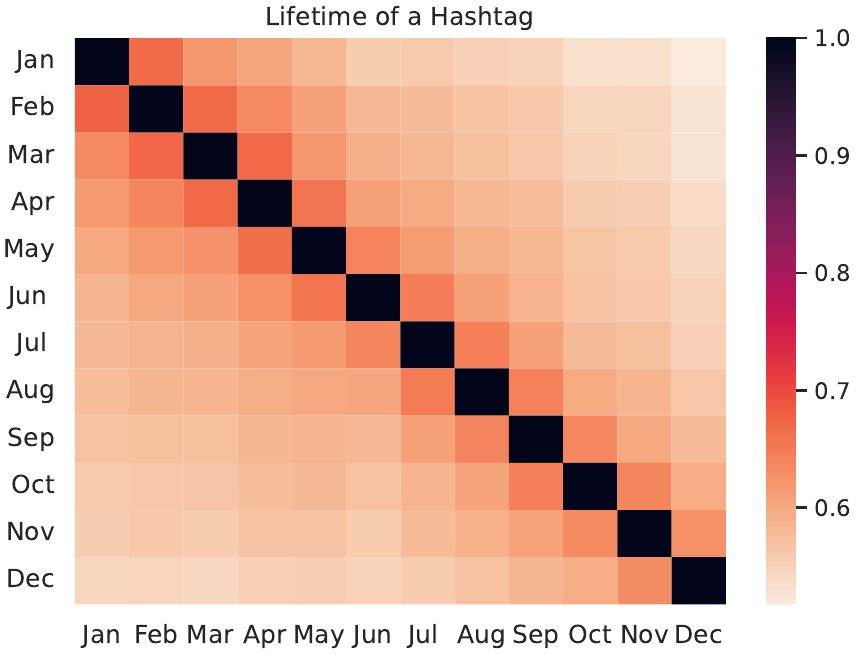}
     \caption{On Twitter, hashtag usage exhibits significant temporal distribution shift, which is challenging for current NLP models. We show the hashtag label set overlap, computed as recall between each month's hashtags, for our longitudinal dataset of 7.13M tweets over the 2021 calendar year. We note that the recall is not symmetric because monthly hashtag sets  have different sizes.}
     \label{fig:overlap}
\end{figure}
%

\begin{figure}[t]
    \centering
     \includegraphics[trim=14cm 8cm 28.5cm 5cm, clip,width=\columnwidth]{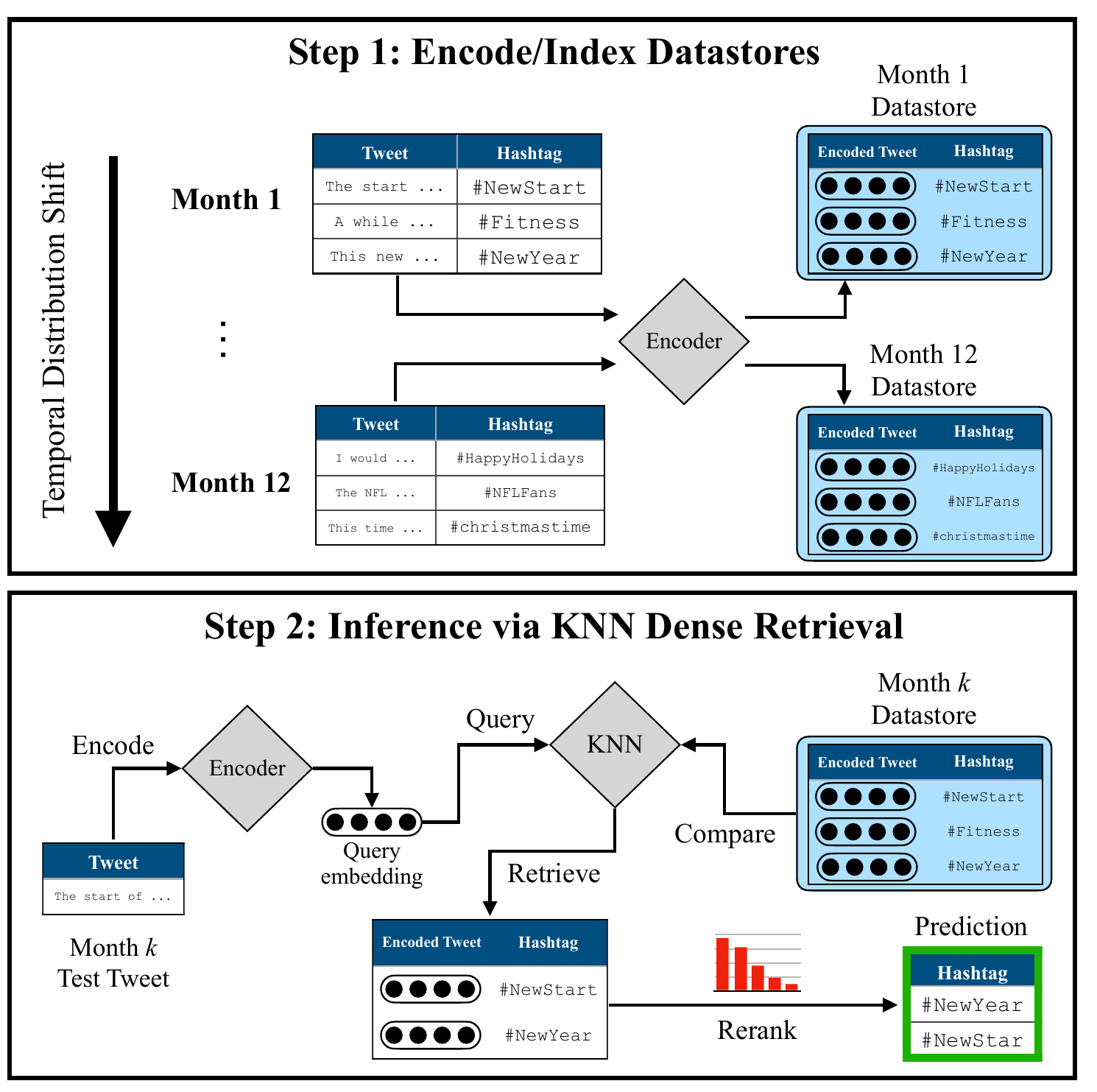}
     \caption{An overview of the KNN Dense Retrieval model components. First, a month's tweets are encoded into embeddings and added to an external, indexed datastore with their associated hashtags---a process that could be automated in a production setting. Second, a test tweet is encoded and used to query the most recent datastore, which contains the latest month's training data for temporal adaptation. The top-$k$ tweets from this datastore are then retrieved and re-ranked.}
     \label{fig:overview}
\end{figure}

In this paper, we perform an empirical analysis of a known, but understudied learning paradigm in the context of temporal adaptation: non-parametric classification via dense retrieval from a datastore. 
By using a $K$-nearest neighbor (KNN) classifier in conjunction with a static neural text encoder (see Figure~\ref{fig:overview}), we demonstrate that continual and automatic updates of the training datastore facilitate simple adaptation and deletion. 
In contrast to contemporary neural  paradigms, which require gradient descent during fine-tuning, our approach (depicted in Figure~\ref{fig:overview}) requires no gradient updates---both adaptation and retroactive deletion are accomplished with minimal compute.

In order to analyze our proposed temporal adaptation method, we introduce a new supervised dataset (\S~\ref{sec:dataset}) focused on \textit{hashtag prediction}, a multi-label classification task in the social media domain that exhibits label shift over time (see Figure~\ref{fig:overlap}). 
Hashtag prediction, in which a system must predict the set of hashtags to be assigned to a given input tweet, is particularly amenable to our non-parametric method due to the intrinsic, user-generated categorization.
In other words, the latest supervised training data can be automatically scraped and updated on-demand.
For the dataset, we collect $7.13$M tweets, binned by week of creation, over the entire 2021 calendar year. 
We explicitly organize this data into twelve non-overlapping time buckets, each of which is sub-divided into weeks, along with train/validation/test splits. 
Subsequently, we scrape the set of tweets that were deleted 5 months after initial collection as a test bed for deletion experiments.
We plan to publicly release this dataset in the form of tweet IDs upon publication.

In addition to comparing several text encoding strategies and re-ranking methods on the dataset (\S~\ref{sec:res}), we compare against two strong, state-of-the-art parametric baselines based on BART~\cite{lewis2019bart}, a large pre-trained sequence-to-sequence model.
These variants include a fine-tuned BART encoder-based classifier, along with the full, unconstrained seq2seq generation model fine-tuned on tweet-hashtag pairs.
We find that our best non-parametric approach outperforms \textit{static} parametric models with an average relative gain of  $64.12\%$ recall when the test distribution shifts---and, even outperforms conventional gradient-based \textit{temporal adaptation} of parametric baselines with an average relative gain of $11.58\%$ recall.
%

%
Together, our empirical analyses highlight non-parametric techniques as a practical and promising direction for adaptation to distribution shift and user-deletion, and may facilitate future work on real-world, temporal NLP system deployment.

\section{Related Work}

\paragraph{Hashtag Prediction.} Hashtag prediction was first studied in the context of predicting which hashtags will go viral in the future~\cite{ma2012will}. 
We consider a related but different task of suggesting a \textit{set of hashtags} for a given tweet. 
More related methods were proposed that compute tf-idf vectors from extracted keywords and apply simple classifiers on these vectors for hashtag recommendation~\cite{jeon2014hashtag,sedhai2014hashtag}. 
\citet{feng2014we} collected their own dataset of millions of tweets/hashtags across two months of 2012, studied temporal hashtag usage distributions over this period, and designed a personal hashtag recommendation model using user metadata features.
Instead, we release a public dataset of tweets over an entire year, study how to continually and automatically update a datastore for the task, and only consider tweet text features that respect user privacy and right-to-deletion laws.
Later methods utilized deep learning architectures such as LSTMs~\cite{li2016tweet,shen2019hashtag}, CNNs~\cite{gong2016hashtag}, and other deep architectures~\cite{ma2019co} for hashtag recommendations.
To the best of our knowledge, this is the first approach to apply dense retrieval for hashtag recommendation.

\paragraph{Temporal Distribution Shift.}

Other early research models semantic shift of the same words as one type of temporal change~\citep{wijaya2011understanding,kulkarni2015statistically,hamilton-etal-2016-diachronic,kutuzov-etal-2018-diachronic}. Separately, \citet{10.1145/1143844.1143859,wang2012continuous} add temporal information into topic models, and ~\citet{huang-paul-2018-examining,he2018time} analyze standard domain adaptation methods for temporal document classification.
More recently, the temporal generalization problem has been re-emphasized in pretrained language models~\citep{lazaridou2021mind,dhingra2022time,luu-etal-2022-time,jin-etal-2022-lifelong} as their re-training cost continues to grow. 
\citet{loureiro2022timelms} adapts language models to diachronic shift by continually updating model parameters through gradient descent on new data at regular time intervals.
The temporal distribution shift problem on Twitter has been studied in~\citet{preotiuc-pietro-cohn-2013-temporal,rijhwani-preotiuc-pietro-2020-temporally,luu-etal-2022-time,kowald2017temporal,kamath2013spatio}. Most of these works explore traditional domain adaptation techniques that require re-training the model, the only exception is~\citet{preotiuc-pietro-cohn-2013-temporal} where they keep track of the latest hashtag frequencies as the prior to adapt a naive Bayes classifier. In this work, we study non-parametric methods that adapt models through dense retrieval without any retraining. 

\paragraph{User Data Deletion.}
With a recent focus on providing users with control over when and how their data is used, many approaches have been proposed to address utilizing ML models when users have revoked the use of their data. 
While several methods have been proposed to partially mitigate computational cost through techniques to speed up re-training, substantial compute is still required~\cite{bourtoule2021machine,wu2020deltagrad}. 
Decremental learning~\cite{cauwenberghs2000incremental,karasuyama2010multiple} has been proposed for removing specific training samples from already learned models while maintaining comparable performance. Later work studied the problem of removing data from $k$-means clustering~\cite{ginart2019making}. Other approaches~\cite{guo2019certified,golatkar2020eternal,pmlr-v132-neel21a} have taken a differential-privacy approach to forgetting user data by formulating it as an obfuscation task such that parameters of a trained model are modified such that user data is non-identifiable.
Different from previous methods, we explore a simple, theoretically guaranteed, and effective way to delete user data in this paper by modifying the datastore.
There are also approximate deletion methods~\cite{nguyen2020variational,Golatkar_2021_CVPR}, which do not actually remove the sample completely, and are out of the scope of this paper.

\section{KNN Dense Retrieval Model}\label{sec:model}

We propose a simple but effective solution to distribution shift and user data deletion based around a non-parametric classifier.
Unlike parametric models which need to be re-trained or fine-tuned when train or test data changes, non-parametric models can be quickly updated simply by replacing or amending the datastore which contains the training data.
We propose using a dense $K$-nearest neighbor (KNN) classifier with a neural text encoder~\cite{altman1992introduction,khandelwal2019generalization}, which retrieves the top $K$ nearest points from training data based on the neural encodings of the samples.
%
The retrieval of the top-$K$ samples works as follows:

\begin{equation}
    \mathcal{Y}_K(x) = \textrm{top-$K$-argmin}_{(x',y') \in \mathcal{D}} \ \ \|e(x) - e(x')\|^2_2, 
\end{equation}

where $e(x)$ is the neural encoding of the test tweet $x$, and $x'$ represents training sample tweets, stored in a datastore. $y'$ is the class label (hashtag) of $x'$. We use L2 as the distance metric. In our experiments, we found that re-ranking the top-$K$ tags, and returning only top $R$ of the re-ranked tweets (where $K>>R$) yields higher recall, so we add a re-ranking step to the KNN retrieval:

\begin{equation}
    \mathcal{\hat Y}_R(x, \mathcal{Y}_K(x)) = \textrm{top-$R$-rerank}(x, \mathcal{Y}_K(x)),
\end{equation}

which re-ranks and returns the top $R$ tags. Based on this, the dense KNN models would have three main components: (1) A static neural encoder (static as in it is only trained once and doesn't need to be updated) (2) a datastore that enables fast nearest neighbor search and (3) a re-ranker. Below we explain these components in more detail.

\subsection{Neural Tweet Encoder}

We use representations generated by a transformer-based model (BART,~\citet{lewis2019bart}) to encode the tweets for saving and retrieval. The encoding is very crucial, as it is the main tool that helps us find relevant hashtags, and we find that if the encoder model is not appropriately trained (on relevant data), it can adversely impact the performance of the model. We empirically ablate different transformer-based encoders with different training objectives in Section~\ref{sec:ablation}. 
It is noteworthy that the encoder is only trained once, on historic training data (time bucket $1$), and it is \textit{not} temporally updated.



\subsection{Prediction Task} 
Users very commonly add more than a single hashtag to a tweet they are composing. In fact, in our dataset, presented in Section~\ref{sec:dataset}, we find that on average users assign nearly three hashtags to each individual tweet. 
Thus, we treat the hashtag prediction task as a \textit{multi-label} classification problem, where each datum has two components: a source tweet text and a target set of its associated hashtags that must be predicted. 
We describe prediction in our datastore setup in the next subsection.

\subsection{Datastore and Search}\label{sec:datastore}

For storing the tweets and hashtags in the datastore as key-value pairs,  we ``unroll'' tweet/hashtag sets. This means that if the tweet is ~\textit{``Happy New Year  \#welcome2021 \#growth \#change \#newyear''}, we unroll it into four separate training samples, each with the tweet text as source, but only a single hashtag as target, and then save each of these four separate pairs as a key-value pair.
%
%

The engine behind many modern non-parametric methods is an efficient indexing and search algorithm that allows for fast retrieval of approximate nearest neighbors~\cite{khandelwal2019generalization,khandelwal2020nearest}.
We follow prior work and use FAISS~\cite{johnson2019billion}, which is a data structure that enables efficient similarity search and clustering of dense vectors based on product quantization. FAISS performs approximate nearest neighbor search over approximate distances due to vector quantization, which is faster than an exact search, but does not precisely correspond to exact L2 distance.
We explain the implementation details of the datastore in Section~\ref{sec:experimental_setup}.


\subsection{Re-ranking Retrieved Hashtags}

%
To re-rank the retrieved $K$ tweets for better performance, we try the three methods below:
%
%

\begin{enumerate}
    \item {\bf Default distance ranking:} In this method, we use the L2 distances returned by FAISS, which are approximate as FAISS quantizes the vectors to speed up the search. We then rank the hashtags from nearest to furthest and return the ``unique'' top $R$, as there might be repetitions in the top retrieved hashtags. 
    
    \item {\bf Actual distance ranking:} This is similar to the previous method, however, instead of using FAISS's approximate distance, we find the actual distance between the encoded query tweet and its retrieved neighbors. We then rank the neighbors based on this new distance and return the top $R$.
    
    \item {\bf Frequency-based ranking:} This method is similar to the conventional KNN classification, but with a multi-label twist, where we count the number of occurrences of each hashtag in the $K$ retrieved tags, and then rank them from most repeated to least, and return the top $R$ most common ones. This method prioritizes the $R$ hashtags that had the greatest support in the initial $K$ retrieval.
\end{enumerate}



\section{Experimental Setup}\label{sec:experimental_setup}

In this section, we introduce our new longitudinal hashtag prediction dataset, and discuss baseline model specification and training, along with the metrics used for evaluation.

\subsection{Longitudinal Hashtag Prediction Dataset}\label{sec:dataset}

To evaluate a model's ability to efficiently adapt to temporal distribution shift in hashtag prediction, we require a dataset with temporal annotations so that data can be bucketed in a fine-grained manner. 
As far as we are aware, no such dataset is publicly available. 

Thus, we collect a novel, public large-scale benchmark dataset for temporal hashtag prediction on Twitter data.\footnote{Code and data will be available upon publication.}
We build our dataset by scraping tweets\footnote{We use the Twitter API at \url{https://developer.twitter.com/en/docs/twitter-api}.} from the entire 2021 calendar year, grouping tweets by the week in which they were published. 
For each week, we only keep tweets that contain at least one hashtag from the top-$10K$ most frequent hashtags that week. 
Further, we drop infrequent hashtags (i.e. hashtags not in the top-$10K$ per week) from the label set. 
This results in large label distribution shift across time in the dataset, which we visualize in terms of hashtag type overlap in Figure~\ref{fig:overlap}. Of course, temporal shift in the contents of the underlying tweets and how tweet contents correlates with hashtag choice is also present, though less visible in the dataset's construction. 
We preprocess the data to partition each week into train/validation/test sections, and balance the number of train/validation/test tweets across weeks. 
Finally, we scrape the set of tweets that were deleted 5 months after initial collection as a test bed for deletion experiments.
The main statistics of the dataset are shown in Table~\ref{tab:stats}.
We run evaluations on this dataset with three different temporal setups, explained in Section~\ref{sec:experimental_setup}.

\begin{table}[t]
    \centering
    \begin{adjustbox}{width=0.75\linewidth, center}
\begin{tabular}{lc}
\toprule 
Statistic & Value\\
\midrule
Number of Train Tweets Per Week &	475437 \\ 
Number of Val-Test Tweets per Week & 59430 \\
Number of Avg Tags  per Tweet & 2.9 \\
Number of Unique Tags per Week & 10000 \\
Avg Hashtag Length (tokens) & 3.2 \\
Avg Tweet Length (tokens) & 30.3 \\
\bottomrule
\end{tabular}

    \end{adjustbox}
    \caption{Summary statistics of our longitudinal Twitter hashtag dataset.}
    \label{tab:stats}
    \end{table}


%
\paragraph{Dataset Bucketing for Temporal Adaptation.}
To facilitate temporal evaluation, we bin the dataset into 12 buckets, each consisting of four weeks (near, but not exactly aligned with month boundaries). In each bucket, the first three weeks are intended to be used as a contemporaneous datastore for temporal adaptation (or as a re-training/fine-tuning set, depending on the method being employed). For static evaluation of models (i.e. without temporal shift) we further break each week into train, validation, and test sections with an 80/10/10\% split. When evaluating in the temporal setting, the test section from the fourth week in each bucket is used as the test set. This ensures that the corresponding datastore / re-training dataset, while being nearly contemporary with the corresponding test set, is realistic in that it does not come from the future. Finally, some models require a frozen historical training corpus (e.g. to train an encoder to be used for dense retrieval). We reserve the first time bucket for this purpose.

\subsection{Evaluation Metrics and Temporal Setups}
\label{sec:eval_set}

We report recall on top-5 and top-1 retrieved hashtags as our main metric. This means we collect the retrieved hashtags from each method, choose the top-5 or top-1, and then compare that against the gold target (i.e. hashtags that appear in the tweet) and report the recall.
We perform our evaluations with three different temporal setups:

\begin{enumerate}
    \item {\bf Non-temporal:} where there is \textit{no} temporal distribution shift between train and test, meaning that the test set is from the same time interval that the training set is from (the test set is the aggregate of test sets from the first three weeks). i.e. it is conventional, non-temporal evaluation. 
    \item  {\bf W/o Adaptation:} a temporal evaluation setup where the model is trained on/datastore is created on  data bucket $1$ (weeks 1-3) and evaluated on the 12 test buckets (weeks 4 through 48).
    \item {\bf W/ Adaptation:} a temporal evaluation setup  where the training and test data come from the same time bucket (e.g., for test week 8, the model is trained on weeks 5-7, and for test week 48, the model is trained on weeks 45-47). 
\end{enumerate}
Both W/ and W/o adaptation setups have temporal distribution shift between train and test. Non-temporal setup has no distribution shift.

\subsection{Baselines}

\paragraph{Neural Classifier.} We use BART-large~\cite{lewis2019bart} with a multi-label classification head as the neural, parametric classifier baseline. Our main baseline is the classifier trained on time bucket $1$'s data, which we use for the `W/o Adaptation' evaluation.
This classifier has $16886$ labels, which is equal to the training hashtag vocabulary size of time bucket $1$. 
For the `W/ Adaptation' setup, we train $12$ classifiers, on the 12 time buckets (explained in Section~\ref{sec:experimental_setup}). We train each classifier for $30$ epochs, and choose the best checkpoint based on validation recall where the validation data overlaps in time with the training data.
We use learning rate of $3e-5$ with a polynomial scheduler and training batch size of $36$.
%
%

\paragraph{Neural Sequence to Sequence Model.}
To provide a baseline that is not  restricted to a pre-set vocabulary (i.e. the neural classifier), we experiment with a sequence-to-sequence model, which is potentially capable of generating hashtags that it has not seen in the training set, as we want to show temporal degradation isn't simply about the fixed label set.
We use BART with a conditional generation head. The sequence-to-sequence model is trained on the tweet text as input and the sequence of concatenated hashtags as the output. 
For decoding (generation), we run inference on the network, and make it generate $120$ tokens, and then select the first $5$ hashtags for calculating recall. The order of generated hashtags does not affect the score.
Similar to the classifier baseline, we run training for $30$ epochs, with learning rate of $3e-5$, polynomial scheduler and training batch size of $36$.
%


\subsection{KNN Implementation Details}

\paragraph{Encoder.} We explore multiple encoder choices for obtaining tweet encodings for dense retrieval, finding that the encodings from the BART classifier (the last hidden state that is fed to the classifier head) provides the best validation performance. 
In this setup, BART is fine-tuned on the hashtag classification task for time bucket $1$'s data and is used for encoding \textit{every} time bucket. 
That is, we \textit{do not} fine-tune a separate encoder for each time bucket. 
We use this method in our main results, and describe other encoder variants and their results in Section~\ref{sec:res}. 

%
%

\paragraph{Datastore.} We encode all the tweets into a float32  Numpy memory map for keys. The dimension of the keys memory map is $N\times E$, where $N$ is the number of tweets in the training set. Given how we unroll tweets (cf.  Section~\ref{sec:datastore}), $N\geq |D|$, $|D|$ is the number of unique tweets in the dataset (size of the dataset). $E$ is the dimension of the encoded vector, $1024$ in our case with the classifier encoder.
%
We also construct an unsigned int16  memory map for the hashtags, named values. We encode the hashtags by feeding them to the BART tokenizer and using the token ids (we do not embed the tags, we only tokenize and encode them using BART's vocabulary) which  has dimension $N\times V$, where $V$ is the length of the hashtag in tokens and is set at the upper bound $280$, since a tweet is maximum $280$ characters long, and each token is at least a single character. We add padding to the hashtags that are shorter than $280$ tokens. 

%
%
We use the \texttt{IndexFlatIP} quantizer along with $L2$ distance as the FAISS metric. We also set the number of keys that are added at a time to be $500k$. For retrieval, we use their \texttt{search} function as well, which retrieves the nearest $K$ neighbors but using the approximate (quantized) distances which means the retrieval and order is not exact. We also experimented with Cosine distance and found it to under-perform compared to L2 in our case.


\begin{table}[t]
\centering
\begin{adjustbox}{width=0.99\linewidth, center}
 \newcolumntype{L}{>{\RaggedLeft\arraybackslash}p{0.06\linewidth}} 
  \newcolumntype{O}{>{\RaggedLeft\arraybackslash}m{0.07\linewidth}} 
  \newcolumntype{D}{>{\arraybackslash}m{0.15\linewidth}} 
  \newcolumntype{R}{>{\arraybackslash}m{0.29\linewidth}} 
\begin{tabular}{@{}lccccccccccccccc@{}}
	\toprule
 	 {} &\multicolumn{2}{c}{}& & \multicolumn{5}{c}{Temporal}	 \\
    \cmidrule{5-9}
	 {\multirow{3}{*}{}} &\multicolumn{2}{c}{Non-temporal}& &\multicolumn{2}{c}{W/o Adaptation}	&   {}& \multicolumn{2}{c}{W/ Adaptation}	 \\
	\cmidrule{2-3} \cmidrule{5-6} \cmidrule{8-9} 
	&                                  {$@$5} &  {$@$1}          &                   & {$@$5} &  {$@$1} &                   & {$@$5} &  {$@$1}  \\
    \midrule
Frequency baseline  &	{1.68}&	{0.47} &&1.11&	0.26&	&1.69&	0.46                                                 \\
Classifier	 &	\textbf{39.87}	& \textbf{14.80} & &14.79&	5.41&	&23.25&	8.34                                                     \\
Seq2Seq	    &	34.36&	12.96 & &15.97&	6.10&	&23.49&	8.30                                                    \\
\textbf{KNN-Clsf}	&	{39.54} &	{13.15}&&\textbf{18.45}&	\textbf{6.65}&	&\textbf{26.21}&	\textbf{9.60}     \\
	\bottomrule
\end{tabular}

\end{adjustbox}
\caption{Summary of performance comparison between different methods, in terms of temporal and non-temporal recall at top-1 and top-5. The frequency baseline returns the most common hashtag from the training set. The KNN-clsf is the KNN model using the BART classifier encodings for the datastore. The three different evaluation setups (Non-temporal, W/o Adaptation and W/adaptation) are explained in Section~\ref{sec:experimental_setup}.}
\label{tab:summary}
\end{table}

\begin{table}[t]
\centering
\resizebox{0.9 \columnwidth}{!}{
\begin{tabular}{@{}llllcccccc}
\toprule 
	K/Method	&20	&50	&100	&1024	&2048 \\
\midrule 
\textbf{Frequency-based}	    &\textbf{25.23}	&\textbf{26.26}	&\textbf{26.21}&	\textbf{23.12} &	\textbf{21.44}\\   
Default Dist.	&24.00	&24.09	&24.32&	24.27   &	24.21\\ 
Actual Dist.	   &24.22	&24.50	&24.77&	24.97 &	24.92\\   
\bottomrule
\end{tabular}}

\caption{Ablation study of the effect of different $K$ for KNN retrieval, and different re-ranking methods. The re-ranking methods are explained in the Section~\ref{sec:model}. Recall $@5$ is reported, averaged over the $12$ test weeks from different time buckets, under the W/ Adaptation evaluation setup.  The three different evaluation setups (Non-temporal, W/o Adaptation and W/ Adaptation) are explained in Section~\ref{sec:experimental_setup}.}
\vspace{-3ex}
\label{tab:ablate_k}
\end{table}

\section{Results}\label{sec:res}

\begin{figure*}[h!]
    \centering
\includegraphics[trim=4cm 0.1cm 4cm 1.5cm,clip,width=0.92\linewidth]{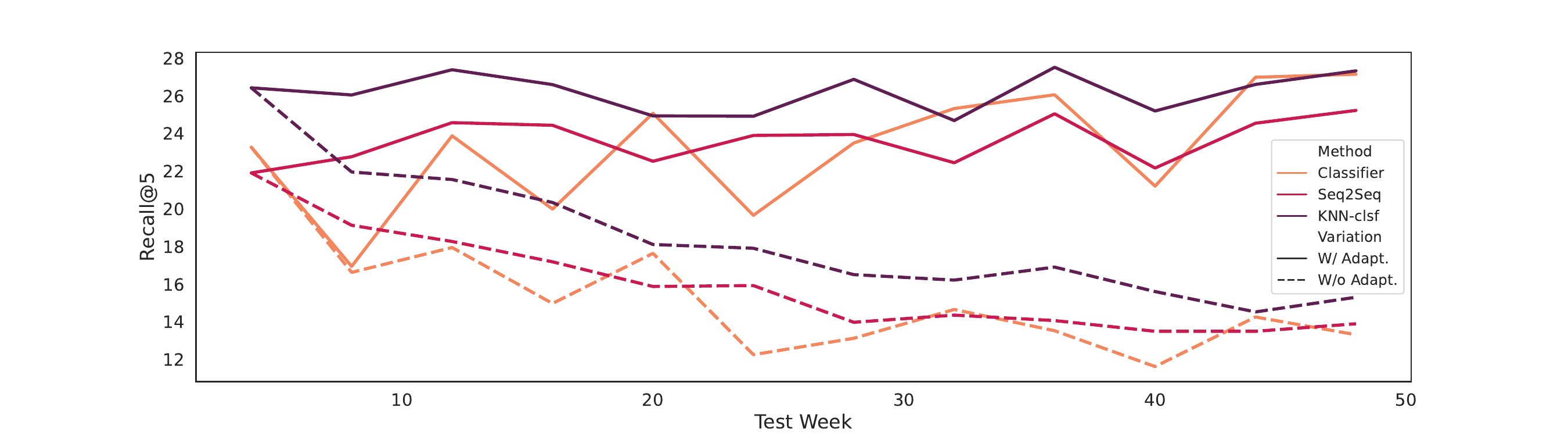}
\vspace{-1.5ex}
     \caption{Comparison of the KNN classifier with the parametric neural sequence-to-sequence and classifier models, in terms of the recall on top 5  hashtags. The solid lines show evaluation results of a model trained/datastore created on the \textbf{time bucket 1} (first 3 weeks) and tested on the test week. The dashed lines, however, show the performance of adapted models, as in models that are trained on the corresponding data bucket.}
     \label{fig:temporal}
    \vspace{-1ex}
\end{figure*}

\begin{figure}[h!]
    \centering
\includegraphics[width=0.9\linewidth]{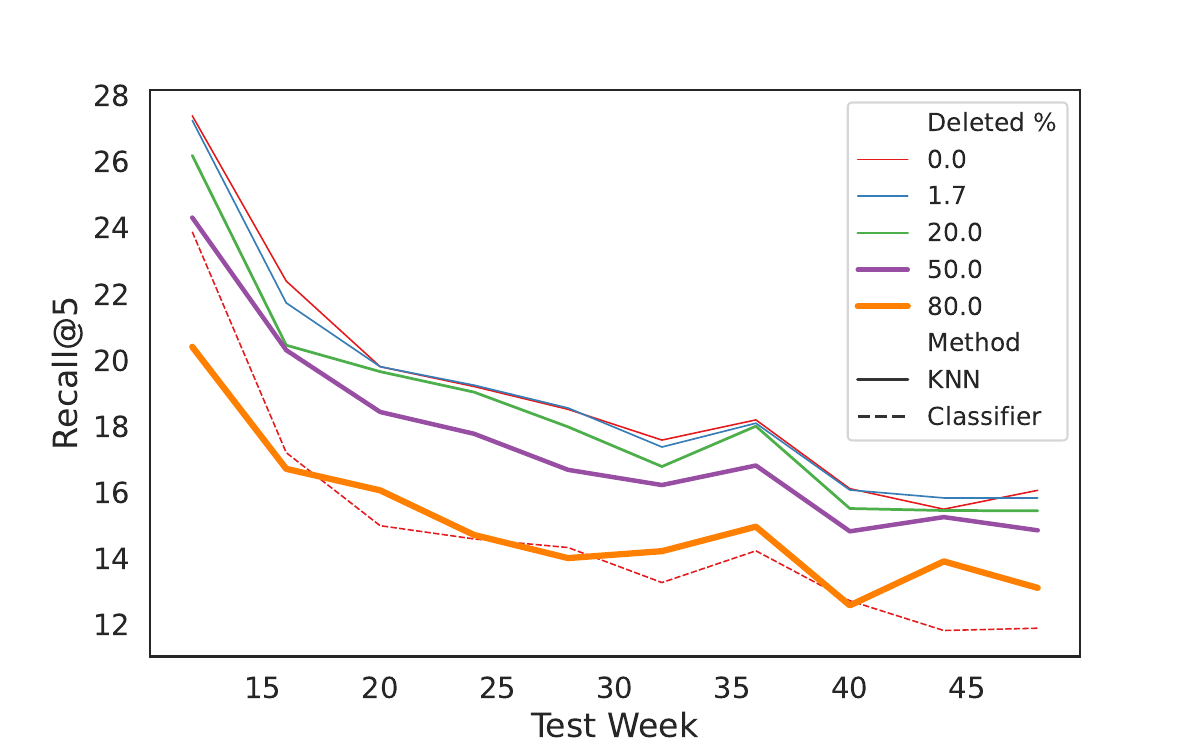}
\vspace{-1ex}
     \caption{Results of the data deletion experiment, where we delete different proportions of the training data in a time bucket and observe how the performance degrades. Thickness of the line has correlation with amount of deleted data. The $1.7\%$ is actually user deleted data, but the rest is randomly deleted. Evaluation follows the temporal W/o Adaptation setup, explained in Section~\ref{sec:experimental_setup}. }
     \label{fig:deletion}
    \vspace{-1ex}
\end{figure}

\begin{table}[t]
\centering
\begin{adjustbox}{width=0.99\linewidth, center}
 \newcolumntype{L}{>{\RaggedLeft\arraybackslash}p{0.06\linewidth}} 
  \newcolumntype{O}{>{\RaggedLeft\arraybackslash}m{0.07\linewidth}} 
  \newcolumntype{D}{>{\arraybackslash}m{0.15\linewidth}} 
  \newcolumntype{R}{>{\arraybackslash}m{0.29\linewidth}} 
\begin{tabular}{@{}llcccccccccccccc@{}}
	\toprule
 	 {} &\multicolumn{2}{c}{}& & \multicolumn{5}{c}{Temporal}	 \\
    \cmidrule{5-9}
	 {\multirow{3}{*}{}}  & \multicolumn{2}{c}{Non-temporal}&&\multicolumn{2}{c}{W/o Adaptation}	&   {}& \multicolumn{2}{c}{W/ Adaptation}	 \\
	\cmidrule{2-3} \cmidrule{5-6} \cmidrule{8-9} 
	&                                  {$@$5} &  {$@$1}          &                   & {$@$5} &  {$@$1} &                   & {$@$5} &  {$@$1}  \\
    \midrule
Frequency baseline	    &1.69&	0.46   &&{1.68}&	{0.47} &&1.11&	0.26                                               \\
KNN-Seq	            &    	     35.21&	11.83&&10.35	&5.88&&	19.64&	3.32 \\
KNN-Bertweet	    &           33.07&	11.00&&7.55	&4.59&&	15.46&	2.30\\
\textbf{KNN-Clsf}	&	\textbf{39.54} &	\textbf{13.15}  &&\textbf{18.45}&	\textbf{6.65}&	&\textbf{26.21}&	\textbf{9.60}\\
	\bottomrule
\end{tabular}

\end{adjustbox}
\caption{Ablation study of the effect of different encoders for encoding the tweets in the datastore and retrieving the nearest neighbors.}
\vspace{-3ex}
\label{tab:enc}
\end{table}


\begin{figure}[h!]
    \centering
\includegraphics[width=0.9\linewidth]{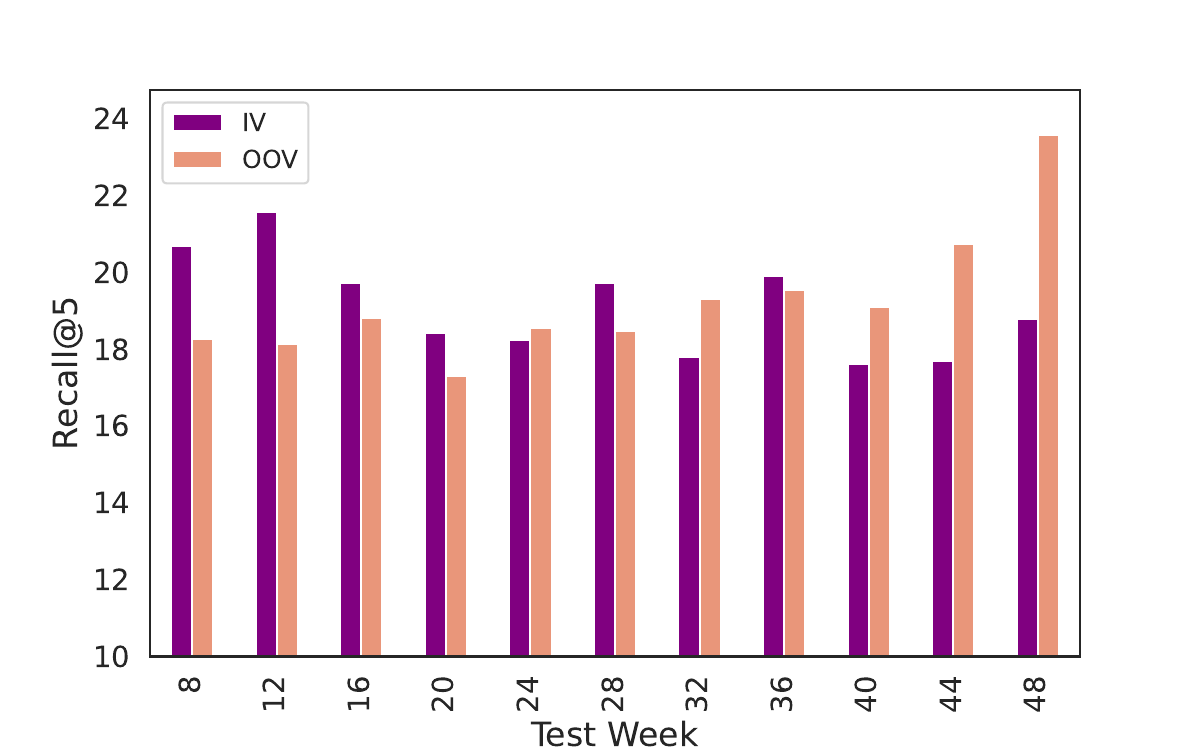}
     \caption{Studying the effect of updating the datastore on capturing  out-of-vocabulary (OOV) hashtags. OOV recall is reported with respect to time bucket 1's training vocabulary.}
     \label{fig:knnoov}
    \vspace{-4ex}
\end{figure}

    


Our experimental results consist of:  (1) temporal performance of different methods, (2) user-deletion performance decay, and (3) ablation of $K$ in KNN, encoding models, and re-ranking methods.

\subsection{Temporal Adaptation Performance}

Table~\ref{tab:summary} summarizes the comparison of different methods averaged over the $12$ time buckets, under three evaluation setups from  Section~\ref{sec:experimental_setup}. The results in the table are averaged over the $12$ time buckets.
Figure~\ref{fig:temporal} shows the breakdown of the numbers in the table, over the $12$ test weeks.  
The frequency-baseline just returns the top most common hashtag of the time bucket's training data.
Note that the results in the non-temporal column are strictly non-temporal, meaning that they use the train/test sets from the same time interval (3 first weeks of the time bucket).  
This is distinct from the leftmost point in Figure~\ref{fig:temporal}, where the fourth week in the time bucket is used for evaluation and thus has a slightly larger degree of temporal change.

The first intuitive observation from Figure~\ref{fig:temporal} is that for the solid line (W/o adaptation), as the test week gets temporally further and further from the train data. 
We can see that the KNN classifier outperforms both the baselines (BART Seq2seq and classifier), whether the evaluation setup is w/o adaptation or adapted. 
In terms of comparing the two parametric baselines, we can see that in the temporal setups (w/ and w/o adaptation) the seq2seq model outperforms the classifier, which could be due to to how it is capable of generating unseen hashtags, unlike the classifier which is bound to a static pre-determined set.
For non-temporal (conventional) evaluation, however, the neural classifier has the best performance. 
We hypothesize that this demonstrates the robustness of KNN to distribution shift (since both the w/o and w/ adaptation settings are temporal). However, for the non-temporal case where the train/test distribution matches the neural classifier is more accurate.

\subsection{User Tweet Deletion}
As explained in the Introduction, efficient model adaptation facilitates selective deletion of training data, which can be an important feature for a variety of reasons. For instance, harmful training examples that may cause spurious outputs during inference can be removed, or fidelity to user-initiated data deletions can be maintained.
In this section, we study deletion in the parametric and non-parametric models.
For the non-parametric KNN model, deletion involves removing the given tweet from the datastore---deleting the entry and re-indexing the datastore---an operation that takes minutes on CPU.
The classifier, however, needs to be fully trained again, without the deleted tweets, which in our case (using a single A100 GPU and training for 40 epochs) takes about a week. 

We use time bucket $3$ for this experiment (i.e. weeks 9-11 are used for training) and we test on weeks $12$ and later ($16$, $20$, ..., $48$). 
In terms of performance drop, we look at 4 different dataset deletion percentages: $1.7\%$, $20\%$, $50\%$ and $80\%$. The $1.7\%$ shows the scenario in which we re-scraped the tweets 5 months after we originally collected them, and removed the tweets that have become unavailable (through deletion or suspension), from the dataset. The other 3 percentages, however, are generated by taking i.i.d samples from the dataset, and dropping them. 
We compare the behavior of the KNN model with that of the parametric classifier w/o any deletion in Figure~\ref{fig:deletion}.
%
We observe that the KNN model gracefully decays as more data is deleted, and even with the deletion, it still outperforms the upper bound classifier model.
%

\subsection{Ablation Studies}\label{sec:ablation}
In this section, we ablate the different components involved in deploying the KNN model: The encoding of the tweets for building the datastore, the $K$ in KNN, and the re-ranking of the retrieved $K$ nearest neighbors. Finally, we break-down the recall $@5$  results of the KNN over different time buckets, to see how well the updated datastore helps capture out-of-vocabulary tags that a datastore/model from the first  time bucket wouldn't have captured.

\paragraph{Ablating encoder for datastore.}
Apart from the classifier encoder used in all previous experiments, we also tried using the seq2seq model as the encoder, and also a generic encoder trained on tweets, named Bertweet~\cite{bertweet}. We compare these encoders in Table~\ref{tab:enc}. We can see that the classifier has the highest performance, then the Seq2Seq, followed by the Bertweet encoder which is the worst. We hypothesize that the poor performance of Bertweet is due to the outdatedness of its training set, which consists of tweets from $2012$ to $2019$, creating a significant  distribution mismatch.
This hints that the encoder for the KNN also requires updates from time to time, however, it doesn't need to be as frequent as  updates for the neural classifier and seq2seq model, and one update per year might be enough (as KNN performance on test week $48$ is still acceptable in Fig~\ref{fig:temporal}).

\paragraph{Ablating $K$ and re-ranking methods.}
Table~\ref{tab:ablate_k} shows the ablation studies, for re-ranking methods and different $K$ values for retrieval. The temporal setup here is W/ adaptation, and the recall $@5$ is averaged over the $12$ test weeks and reported.
Overall, the frequency-based method outperforms the distance-based methods by a large margin, which we hypothesize is due to the robustness added by the repetitions in hashtags. 
We can see that for the distance-based methods, the overall trend is that higher $K$ is better ($1024$ is on average the sweet spot), however, going even higher doesn't degrade the performance meaningfully. The frequency-based re-ranking, however, degrades significantly if the number of retrieved neighbors is large ($1024$ and $2048)$, which is expected, as more irrelevant but common hashtags are suggested. We get closer to a random frequency-based classifier when $K$ approaches the datastore size.

\paragraph{OOV tag prediction performance.}
Finally, we want to investigate how updating the datastore helps us capture new, out-of-vocabulary (OOV) hashtags, that would not be predicted if we kept using the datastore from time bucket $1$. 
Figure~\ref{fig:knnoov} shows the results for this experiment, where the OOV refers to out-of-vocabulary with respect to bucket $1$'s hashtag vocabulary.
IV refers to in-vocabulary hashtags, which means the tags that appear both in the given test week, and the train data of time bucket $1$. 
We report the recall over the IV and OOV tags separately.
We can see that the updates in the datastores help predict $19\%$ of the hashtags that would otherwise not be predicted, on average across the test weeks.
We see that as we proceed with time buckets, the OOV recall grows, eventually overtaking the IV recall. It is worth noting that the number of IV tags is substantially smaller in later time buckets. We suspect the superior OOV performance on the later time buckets is related to content shift, i.e., the meanings of IV tags may have shifted by the later time buckets.

\section{Conclusion}

In this paper, we study the task of temporal adaptation for hashtag prediction, by introducing a new benchmark dataset consisting of tweets and hashtags scraped over a year.
We then evaluate two neural parametric models (classifier and sequence to sequence) and a non-parametric KNN model on the benchmark dataset.
We show that a simple KNN model that retrieves hashtags from a datastore outperforms the sequence to sequence and classifier baselines when evaluated on test sets that have temporal distribution shift.
We also demonstrate that the KNN model is more suitable for removing deleted samples from the model, which is a common scenario in social media.

\section*{Limitations}
Our proposed dense retrieval method relies on a collection of ``historic data'' which is used to train an encoding/embedding model, to produce the representations that would be used as keys in the datastore. As the KNN method relies on these representations for finding neighbors, it is crucial to train the embeddings on in-domain historic data, that would also be unlikely to receive deletion requests. 
However, as we show in our experiments, this data doesn't go stale for even a year out, and still performs well and provided appropriate embeddings, as the performance of the KNN method does not degrade with the passage of time (as seen in Figure~\ref{fig:temporal}).



\section*{Ethics Statement}

We have made sure that we abide by the Twitter develepment and data usage agreement (~\url{https://developer.twitter.com/en/developer-terms/agreement-and-policy}) for the collection and usage of the Twitter data. Also, upon release of the dataset, we will only make the Tweet IDs available for data re-hydration (and not the user ID or the Tweet body), to  protect the privacy of the users, which is inline with the overall goal of this paper as well.


\bibliography{acl2023}
\bibliographystyle{acl_natbib}

\clearpage
\appendix
\section{Hardware and Software Specifications}

 We use  Huggingface Transformers $4.10$, PyTorch $1.9.1$ with Cuda $10.2$,  and Python 3.8. 
We run our inference/retrieval experiments on a single RTX2080 GPU. We ran our training of the baselines on 4 A6000 GPUS for an overall of 2 week worth of GPU hours.
%
%



\end{document}